\documentclass{svproc}
\usepackage{url}

\usepackage{microtype} % microtypography
\usepackage{booktabs}  % tables
\usepackage{graphicx}
\usepackage{subcaption} 
\usepackage[square,sort,comma,numbers]{natbib}
\usepackage[dvipsnames]{xcolor}

\begin{document}
\mainmatter              % start of a contribution
\title{A Qualitative Review of GenAI-Based Methods for Data Generation and Augmentation in Industrial Computer Vision Applications}
\titlerunning{Qualitative Review of GenAI in Data Generation and Augmentation}  % abbreviated title (for running head)
\author{Paul Koch\inst{1} \and Paul Hofmann\inst{1,2} \and
Ferdinand Waßelewsky\inst{1} \and Adem Karakurt\inst{1} \and Andre Sérs\inst{1} \and Jörg Krüger\inst{1,3}}

\authorrunning{Koch et al.} % abbreviated author list (for running head)
\tocauthor{Paul Koch, Paul Hofmann, Ferdinand Waßelewsky, Adem Karakurt, Andre Sérs and Jörg Krüger}
\institute{
Fraunhofer IPK\\
pascalstraße 8-9, Berlin,Germany \\
\email{paul.koch@ipk.fraunhofer.de} \\
\and
Hamburg University of Applied Sciences (HAW Hamburg) \\
Stiftstraße 69, Hamburg, Germany \\
\and 
Technical University Berlin (Tu-Berlin) \\
pascalstraße 8-9, Berlin, Germany \\
}

\maketitle              % typeset the title of the contribution

\begin{abstract}
AI-driven computer vision applications require a profound database to ensure predictable behaviors and performance. Such predictable behaviors are especially important for industrial applications in gaining trust from users. However, such a database is not readily available in industrial applications, and its acquisition is not trivial either. Active learning methods can be applied to ramp up data within a project deployment to iteratively increase the database, and thus the application predictability. Unfortunately, we observe that this often leads to a loss of user trust in the application, which is difficult to regain once lost. This leads to a “chicken-and-egg” dilemma in which neither the database nor the application is developed. In this work, we review state-of-the-art methods and approaches to further boost the database the initial active data ramp-up phase. Here, we focus on recent advancements in GenAI-based data generation and augmentation methods and review their adaptability on an industrial computer vision classification use case. Although we observe a potential for automatic data ramp-up, we also see a domain miss match in between the source (training environment) and target (industrial use-case) -- regarding context defined in natural language and object characteristics. 
%The abstract should summarize the contents of the paper
%using at least 150 and at most 200 words. It will be set in 9-point
%font size and be inset 1.0 cm from the right and left margins.
%There will be two blank lines before and after the Abstract. \dots
% We would like to encourage you to list your keywords within
% the abstract section using the \keywords{...} command.
\end{abstract}
\section{Introduction}
AI-driven computer vision applications have become increasingly relevant in addressing complex industrial challenges. One such application is image-based classification of unlabeled objects, which is crucial for tasks such as parts sorting, warehouse management, and reverse logistics~\cite{ReviewCVinIndustry}. This problem is particularly pronounced in reverse logistics, where the visual representation of objects within each class can be highly diverse, complicating the classification process~\cite{MVIP}. AI assistance in this domain offers significant potential, enabling automated classification to guide human-based sorting, particularly in the remanufacturing processes of old parts~\cite{EIBASorting}.

To realize the full potential of data-driven AI solutions within industrial computer vision, a diverse, extensive, and well-curated database is paramount~\cite{GANsIndustry}. However, the initiation of industrial projects often faces a “chicken-and-egg” dilemma, where the absence of a comprehensive dataset hinders model development and vise versa~\cite{EIBAActiveLearning}. To overcome these challenges, strategies such as ramp-up and active data collection are employed to incrementally increase the database throughout the project deployment~\cite{EIBAIncrementalLearning, EIBAActiveLearning}. Alongside these strategies, advances in synthetic data generation, particularly using CAD models, are explored to supplement data needs~\cite{SynthDataScrews}.

Despite the strides made in synthetic data generation, a significant domain gap persists when transitioning from simulation to production environments. This gap can lead to unexpected behavior of the model~\cite{RevdiwDomainGap}. Techniques such as domain randomization and style transfer are employed to mitigate these discrepancies~\cite{SynthDataScrews, GAN_image2image, FarmingGans}. Recently, generative AI (GenAI) has emerged as a promising tool for realistic and controllable data augmentation. Using text prompts to modify or generate images, generative AI enriches the diversity of training data, potentially bridging the domain gap~\cite{GDAReviewSurvey, GANsIndustry, GenAIApplicaitonsReview, GenerativeAIinmanufacturing, GenAiinIMLReview, SurverOverviewDA}.

Traditional image augmentation methods, such as altering colors or transformations such as flips and rotations, have been successfully used to increase the representation distribution within datasets. However, these methods are limited in altering the visual fidelity and often retain the core semantics and characterization features of the original images, ensuring consistency with annotations in supervised learning.

GenAI offers the capability to produce large and diverse training data distributions using image-to-image, text-to-image, and image-and-text-to-image generation techniques. Despite their potential, these methods are heuristic by design and can cause deviations from the intended annotations, posing challenges in supervised learning scenarios, especially when data are generated autonomously without human review.

In this paper, we conduct a qualitative review of three relevant GenAI categories in the context of a real industrial computer vision classification application: 1) image personalization, 2) image augmentation, and 3) image synthesis. We discuss the generated images on the basis of their semantics, classification characteristics, and the influence of underlying text prompts. Through this qualitative investigation, our aim is to elucidate the current potentials and limitations of these methods, directing their further development towards addressing the challenges of real-world industrial applications.

\section{Related Work}

\noindent\textbf{GenAI-based Image Augmentation:} Generative AI-based image augmentation can be divided into: A) Style Transfer, which bridges the domain gap between source and target environments, and B) Stable-Diffusion-based Image Denoising, which uses diffusion models to diversify image representations through textual or mask-based prompts while preserving object semantics.

\textit{Textual Guided Denoising} methods like DA-Fusion~\cite{DAFusion} and DIAGen~\cite{DIAGen} use textual input to guide image augmentation, enhancing semantic diversity. Autonomous pipelines further automate the augmentation process~\cite{AutonomousDiversifyAugmentation, DiversifyAugmentation}. \textit{Semantic Guided Denoising techniques}, such as SDEdit~\cite{SDEdit} and autonomous background enhancement~\cite{BackgroundAugmentation}, employ semantic masks and textual inputs to diversify images while maintaining object fidelity~\cite{Semantic_Guided}. 

However, classical image augmentation modifies visual aspects without changing semantics, ensuring consistency with original annotations. Generative AI methods, such as Perfusion, can significantly alter semantics, potentially deviating from original annotations while expanding the representation space of the dataset~\cite{DAFusion}.\\\\
\noindent\textbf{GenAI-based Image Personalization:} Image personalization finetunes pretrained models to associate text-based tokens with relevant images, enabling the generation and diversification of images based on text prompts. Perfusion~\cite{Perfusion} exemplifies this by learning a concept (e.g., “cat”) from images and personalizing new ones based on this concept. 
The precision in defining the context is crucial~\cite{TargetConceptMennerPersonalization}. IDAdapter~\cite{IDAdapterPersonalization} and Dreambooth~\cite{Dreambooth} enhance diversity and identity preservation, although challenges such as overfitting~\cite{OverfittingPersonalization} remain. Methods like HyperDreamBooth~\cite{HyperDreamBooth} and InstantBooth~\cite{InstantBooth} optimize training speed, making personalization processes more efficient~\cite{FastPersonalization}.\\\\
\noindent\textbf{Sim-based Synthetic Images:} Simulation-based synthetic image generation is a traditional approach to augmenting training databases. Unlike generative AI methods, which create novel images directly from input datasets, simulation-based methods use environments to render new representations from various perspectives~\cite{SynthDataScrews}. These methods often rely on object CAD, textures, and simulation design. Although CAD data may be available in industrial contexts, textures are typically not, posing limitations. Moreover, CAD data sensitivity restricts sharing with third parties, especially in reverse logistics scenarios~\cite{EIBASorting}. Nvidia-Diffrec~\cite{nvdiffrecmc}, an AI-based method, can synthesize and scan objects, facilitating simulation-based image synthesis for training downstream applications.

\section{Methods}
\begin{figure}
    \centering
    \includegraphics[width=1.0\linewidth]{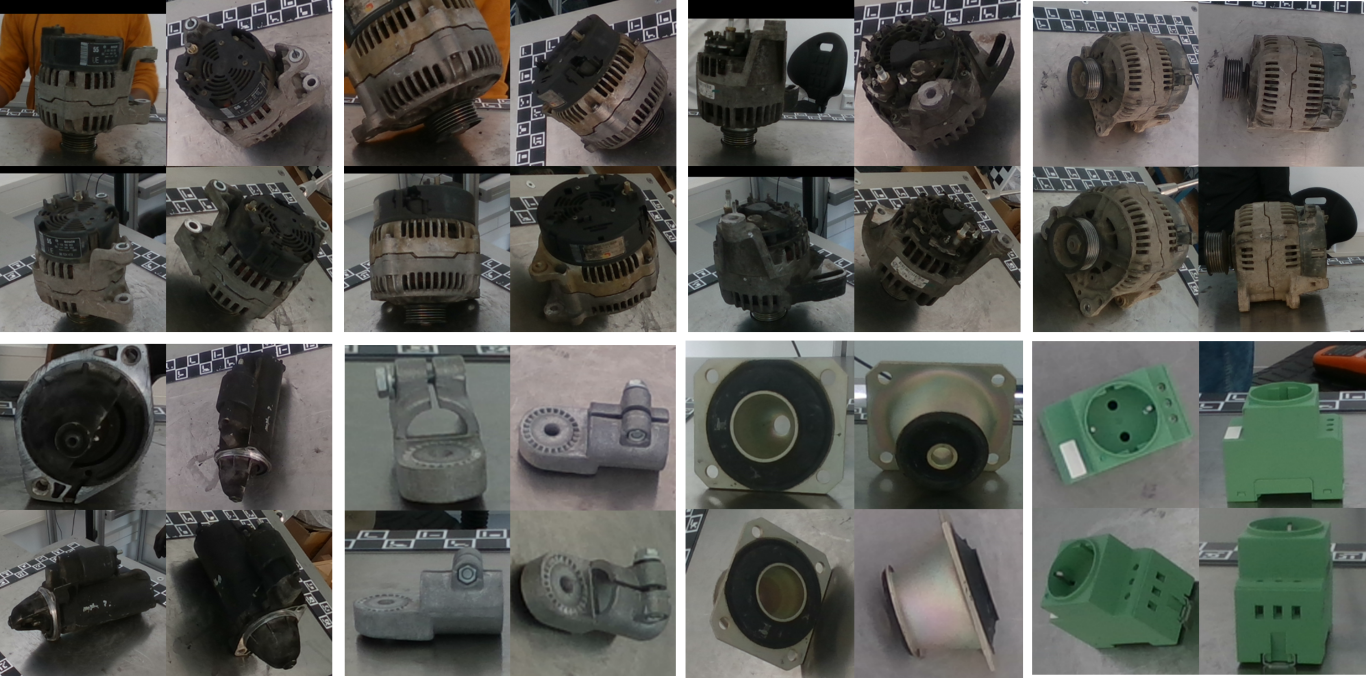}
    \caption{\textbf{MVIP Objects:} The top row features four object classes from the MVIP “generator” super-class. The first three are among the available objects used in our study. The bottom row highlights the object diversity featured in MVIP, where we can include the second object.}
    \label{fig:mvipObjects}
\end{figure}
\textbf{Industrial Use-Case Application and Selected Methods:} In our use-case review study, we utilize the MVIP dataset~\cite{MVIP}, which is specifically designed for the recognition of industrial parts. This dataset comprises 308 industrial object-centered classes, captured from multiple perspectives within an industrial classification context. In particular, MVIP includes a diverse array of used car components, such as starters and generators, sourced from a reverse economy background. These parts possess unique histories that influence their visual representation, making it crucial to rapidly learn from a limited number of class instances to generalize effectively to unseen representations (see Fig.~\ref{fig:mvipObjects}). Additionally, MVIP encompasses numerous object classes that appear visually similar, which poses significant challenges even for skilled human workers. This challenge is particularly pertinent to industrial use-cases, necessitating high image fidelity, a fundamental requirement and challenge for GenAI-based data generation and augmentation. Furthermore, MVIP features natural language tags that describe and categorize objects into task-relevant superclasses, such as “dirty”, “shiny”, “edgy”, “round”, or “rusty” objects. These tags facilitate the exploration of GenAI methods in learning industrial-centric concepts, which may differ from their predominantly human-centric training bases. 
\\\\
\noindent\textbf{Selected Methods and Evaluation:} We assess the utility and adaptability of the methods below within the MVIP use-case in terms of classification accuracy, which challenges both high semantic and classification-specific fidelity due to high object similarities in MVIP. Additionally, we evaluated confidence sensitivity on a generated out-of-distribution (OoD) dataset that is visually similar to MVIP. This analysis aims to determine how the generated images influence AI confidence, a critical factor for predictability and user trust in industrial deployments. Selected Methods:

\textbf{Text-based Image Personalization:} For personalization methods, we select Perfusion~\cite{Perfusion} for our investigations due to its state-of-the-art performance on related benchmarks. Perfusion is noted for its rapid training, high image fidelity, and resistance to overfitting, with image fidelity being particularly crucial to ensure effective classification among visually similar object classes.  

\textbf{GenAI-based Image Augmentation:} For image augmentation, we choose DA-Fusion~\cite{DAFusion} because it is developed with a focus on industrial real-world applications rather than human-centric tasks such as face swapping. DA-Fusion employs text-prompts and eliminates the need for masking.  

\textbf{Synthesizing GenAI for Sim-based Synthetic Images:} Lastly, we investigate NVDIFFRECMC~\cite{nvdiffrecmc}, an enhancement of Nvidia-Diffrec~\cite{nvdiffrec}. This method provides state-of-the-art CAD, texture, material, and lighting generation, which can be seamlessly integrated into any sim-based image generation pipeline.
\\\\
\noindent\textbf{Training Datasets:} In our investigations on MVIP, we utilize a subset of available objects that are readily available to us (see Fig.~\ref{fig:mvipObjects}. We add our generated images with the MVIP images accordingly and conduct training on all 308 MVIP classes. Using the entire MVIP dataset improves the classification challenge and mitigates statistical bias and the impact of random variations. We measure performance across the entire MVIP dataset and individually for the available objects to evaluate the utility of the generated data.

\subsection{Text-based Image Personalization} % Text-to-image personalization? --> Eher nicht, da Bild nicht erzeugt, sondern nur augmentiert wird!
% perfusion

\begin{figure}
    \centering
    \includegraphics[width=1.0\linewidth]{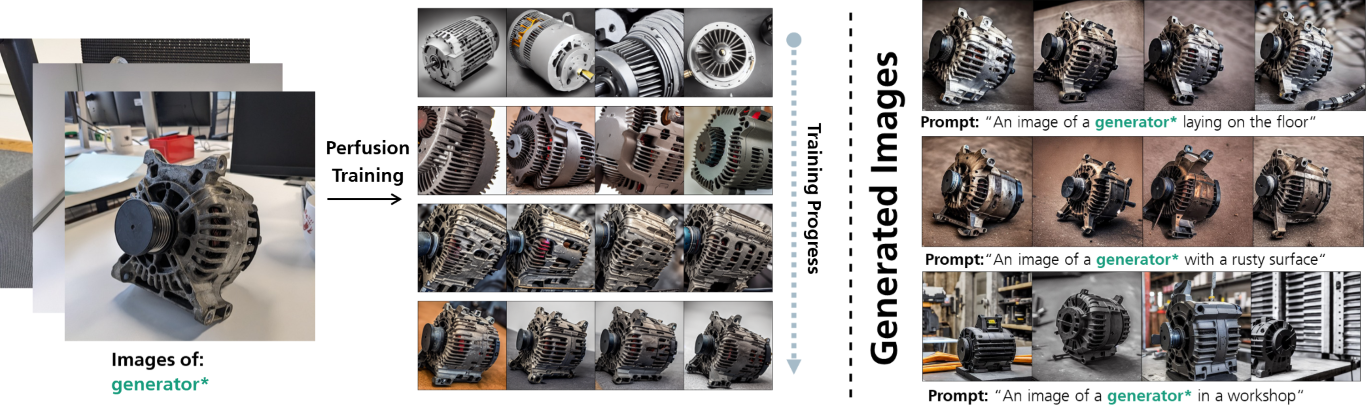}
    \caption{\textbf{Learning new object concepts:} Qualitative Results of Perfusion for generating novel images of objects given a learned new object concept (generator*).}
    \label{fig:perfusion_object_results}
\end{figure}
\begin{figure}
    \centering
    \includegraphics[width=1.0\linewidth]{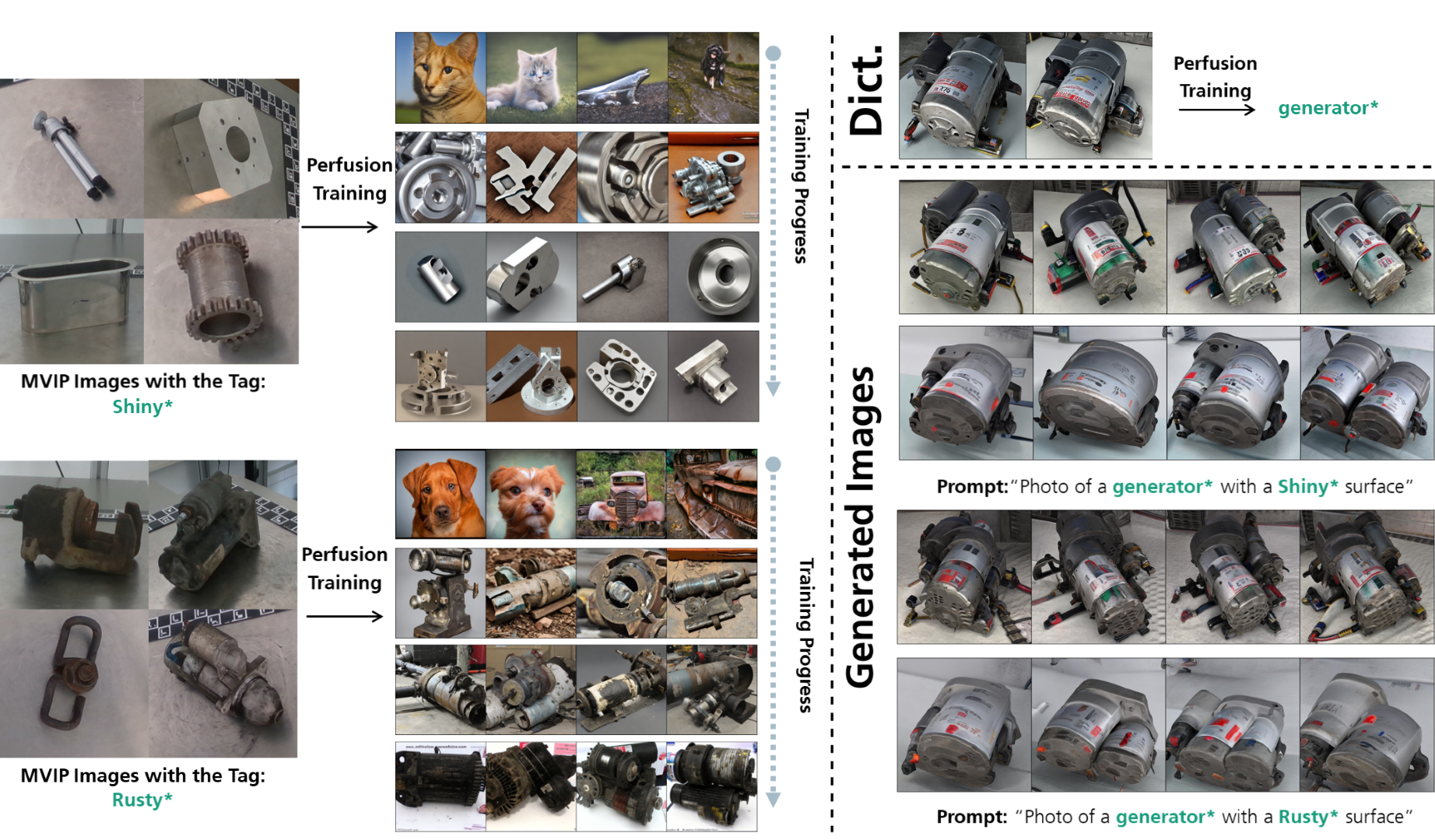}
    \caption{\textbf{Learning new descriptive concepts:} Qualitative results of Perfusion for generating new images of objects given a learned descriptive concept from MVIP~\cite{MVIP} (Shiny* and Rusty*).}\label{fig:perfusion_concepts_results}
\end{figure}

fh
We use Perfusion~\cite{Perfusion} text-based image personalization to finetune its image generation to learn the available concepts of textual objects. Each concept is finetuned individually, thus we can use the MVIP super-classes of e.g. “motor” or “generator” to properly name the resampled object. Using random or ambiguous names (e.g., metal) produces faulty object masking in the perfusion pipeline, which in turn corrects the outcome. We finetune with four images per object, which is yielding the best results in our opinion. Adding more images has no benefit effect. However, we notice that we have to keep a relatively constant perspective on the object. Thus, if one desires to include multiple object sides, one has to train finetune for multiple object concepts per side, e.g. “motor-left”, “motor-right”, “motor-Top”. When fine-tuning perfusion, one has to provide the textual description of the object in a sentence or context: “an image of a generator*”. Using other formulations such as “image of a generator*”, or “generator* shown in a photo” provide much worse results, which is indicating a high sensitivity to formulation. Similarly, we also observe worse results when we are using MVIP`s textual object tags (old, dirty) to generate more detailed image descriptions of e.g “an image of an \textit{old} and \textit{dirty} generator*”,  This makes it hard to integrate such a method in an automated data generation pipeline with robust and consistent outputs. We finetune perfusion for 1000 steps, to avoid overfitting which start to occur afterwards. Most instances peak after 600 steps in their lowest loss. Afterwards, we can prompt the finetuned perfusion models to generate novel images with meaningful queries such as: “an image of a generator* lying on the floor“,“an image of a generator* with a rusty surface“, “an image of a generator* in a workshop“. We show results of these prompts in Fig.~\ref{fig:perfusion_object_results}. 
\begin{figure}
    \centering
    \includegraphics[width=1.0\linewidth]{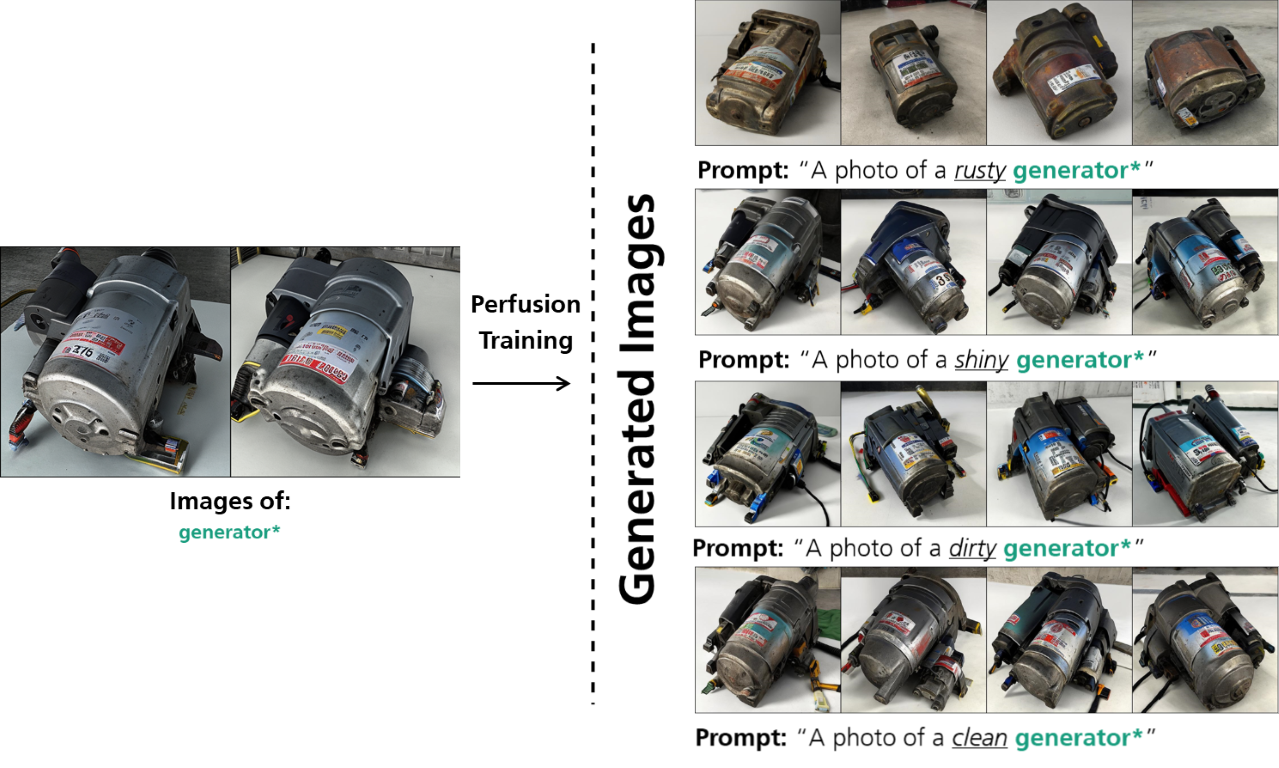}
    \caption{\textbf{Apply known descriptive concepts:} Qualitative Results of Perfusion for generating novel images of objects given subjective and ambiguous descriptive concept from human-centric world knowledge such as “rusty”, “shiny”, “dirty”, “clean”, which are also part of MVIP~\cite{MVIP}.}\label{fig:perfusion_adjective_results}
\end{figure}
To change the appearance of the object, we can add meaningful adjectives (rust, shiny, dirty, clean) in front of the object name, which we visualize in Fig.~\ref{fig:perfusion_adjective_results}. These adjectives are human-centric and, in the context of industrial applications, often ambiguous, e.g., what distinguishes rusty from dirty? Industry-specific concepts are often only known to specialists. Therefore, we would like to further finetune our perfusion base model to also include our adjective concepts. We present the results of such an adjective concept pre-finetuning in Fig.~\ref{fig:perfusion_concepts_results}. 
Here, one has to notice that perfusion starts to learn to map from human-centric concepts (cats \& dogs) to our adjectives (shiny \& rusty), which are valid human-centric animal names for white furred cat or red furred dog. However, we notice a relatively large drop in object fidelity when chaining these human-centric adjectives to learned object concepts. This indicates that we model does fail to properly translate human-centric to industry-centric adjectives. This result is also affected by the dependence of perfusion on the relatively constant perspective mentioned above.       
\\\\
\textbf{Discussion on Image Personalization:} Perfusion yields image or object fidelity. However, with respect to the object classification challenged raised by MVIP (see the top row in Fig.~\ref{fig:mvipObjects}) the fidelity appears not sufficient and would result in corrupting the connect image class annotation. However, Perfusion does yield a controllable image generation pipeline, which can augment the visual distribution with respect to a given task at hand. This is especially interesting for unsupervised~\cite{UnContrast1, UnContrast2, UnContrast3, UnContrast4}, or self-supervised pretraining~\cite{clip, MAE, DINOV1, DINOv2} of the image encoder. Thus, one can use only the original data to finetune the final fully connected classification layer on top of the frozen encoder backbone. With perfusion, one can create a large diverse database for these pretraining methods to yield generalized feature extraction.
% image of a rusty, in a workshop
%
\subsection{GenAI-based Image Augmentation}
% Da-Fusion

In this section, we explore the use of DA-Fusion~\cite{DAFusion} to augment existing images from the MVIP dataset. Unlike Perfusion, DA-Fusion does not generate new images but instead modifies existing ones using latent-space diffusion~\cite{StableDiffusion}. The augmentation process is controlled by an intensity factor which determines the level of noise applied to the input image in latent space, consequently affecting the number of denoising steps during the diffusion process. By adjusting the intensity factor, we can control the degree to which the original image is altered.

With a low intensity factor, DA-Fusion produces augmented images that closely resemble the input images, maintaining high fidelity (see Fig.~\ref{fig:DAfusionslight}). In contrast, increasing the intensity factor results in more significant alterations in the original image (see Fig.~\ref{fig:DAfusionsheavy}). It is important to note that heavily augmented images are unsuitable for supervised learning due to ambiguous fidelity, which can blur classification boundaries between similar objects (see MVIP's similar object problem in the top row in Fig.~\ref{fig:mvipObjects}).

\textbf{Discussion on Image Augmentation:} The slightly augmented images can be incorporated as image augmentation into the labeled training data for supervised learning. The minor changes of small details should not affect the training negatively, but contribute to further generalization which ignores minor artifacts, and thus overfitting. Despite the blurred classification boundary challenges, the heavily altered images remain close to the original feature space, making them valuable as “near” out-of-distribution (near-OoD) data. This proximity to the original distribution makes them useful for unsupervised~\cite{UnContrast1, UnContrast2, UnContrast3, UnContrast4} and self-supervised pretraining~\cite{clip, MAE, DINOV1, DINOv2}. Moreover, near-OoD data are particularly beneficial for supervised contrastive learning~\cite{SuperContrast}, which leverages vague object labels to ensure that related object instances are closely positioned within the pretrained encoder's latent space. This approach should facilitate enhanced feature extraction and thus classification performance~\cite{SuperContrast}. Additionally, DA-Fusion supports object masking to guide the diffusion. This in turn allows the user to control where augmentations are applied, facilitating the user to automatically place and annotate anomalies or defects for anomaly or defect detection applications.   

\begin{figure}
\centering
\begin{subfigure}[t]{0.49\textwidth}
    \centering
    \includegraphics[width=0.9\linewidth]{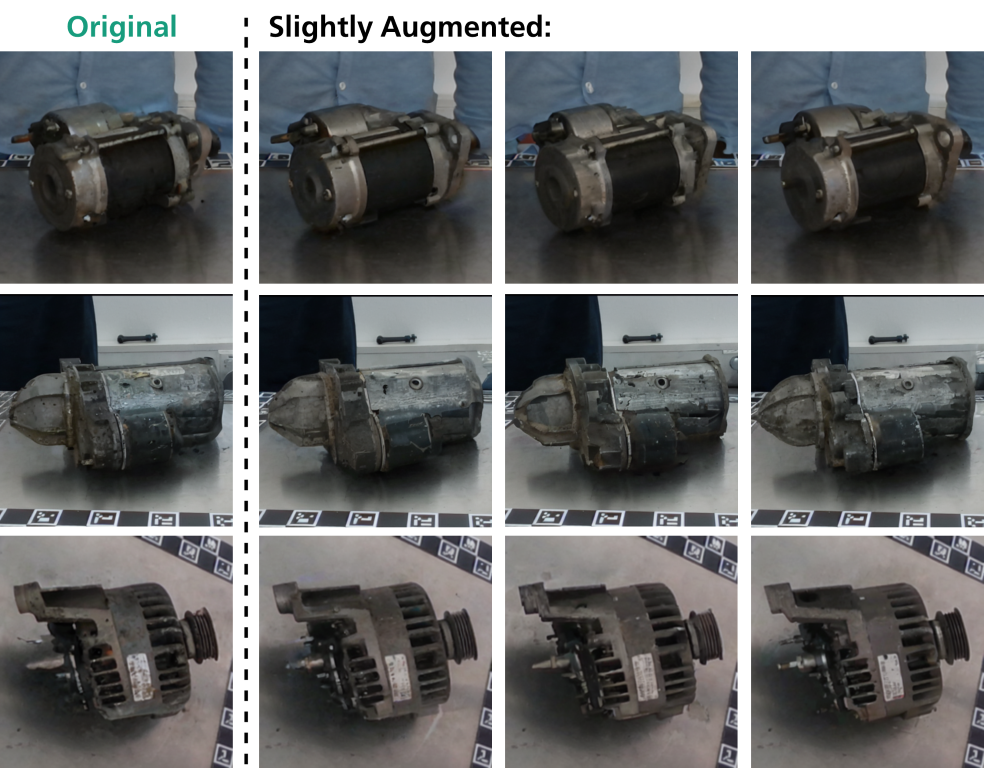}
    \caption{\textbf{DA-Fusion Slight Augmentation}}
    \label{fig:DAfusionslight}
\end{subfigure}
\hfill
\begin{subfigure}[t]{0.49\textwidth}
    \centering
    \includegraphics[width=0.9\linewidth]{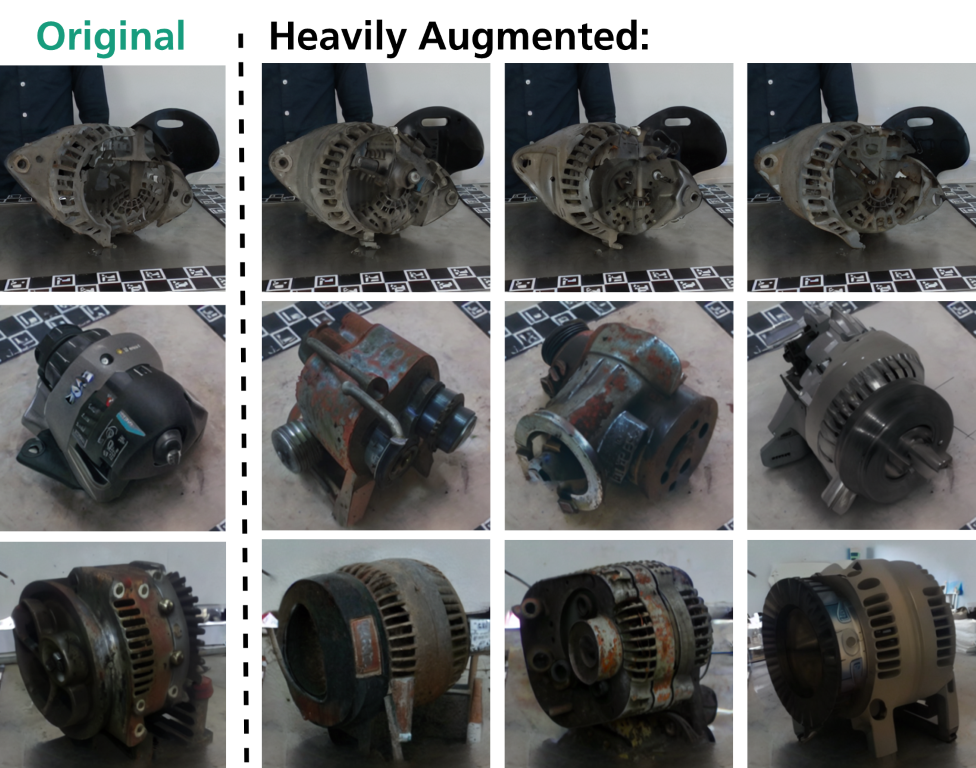}
    \caption{\textbf{DA-Fusion Heavy Augmentation} }
    \label{fig:DAfusionsheavy}
\end{subfigure}
\caption{A) We are using DA-Fusion with 20~\% intensity of diffusion (diffusion strength). The resulting augmented images have a high visual fidelity with minor divination from the original image. The images are mostly alike to the original with minor changes in the texture and semantics. B)We are using DA-Fusion with 50\% intensity. The resulting augmented images have a relatively lower visual fidelity with larger divination from the original image with respect to the slightly augmented images (see Fig.~\ref{fig:DAfusionslight}. The images are mostly alike to the original, but with severe changes in the texture and semantics, which might affect the fidelity in a classification task with highly alike object classes such as MVIP~\cite{MVIP}}
\end{figure}

\subsection{Synthesizing GenAI for CAD-Based Synthetic Data Generation}
%nvidia diffrec... (andre)
This section explores the use of GenAI (NVDIFFRECMV~\cite{nvdiffrecmc}) to synthesize real objects into CAD, texture, and surface assets. These assets are essential for simulations that require rendering photorealistic images from novel perspectives within various scenes. Unlike other scanning methods that primarily focus on CAD generation, NVDIFFRECMV directly infers texture and surface normals from an input image array. This capability enables automation of a simulation-based data generation pipeline without additional engineering, significantly reducing the costs for scaling in industrial applications.

\begin{figure}
\centering
\begin{subfigure}[b]{0.49\textwidth}
    \centering
    \includegraphics[width=0.9\linewidth]{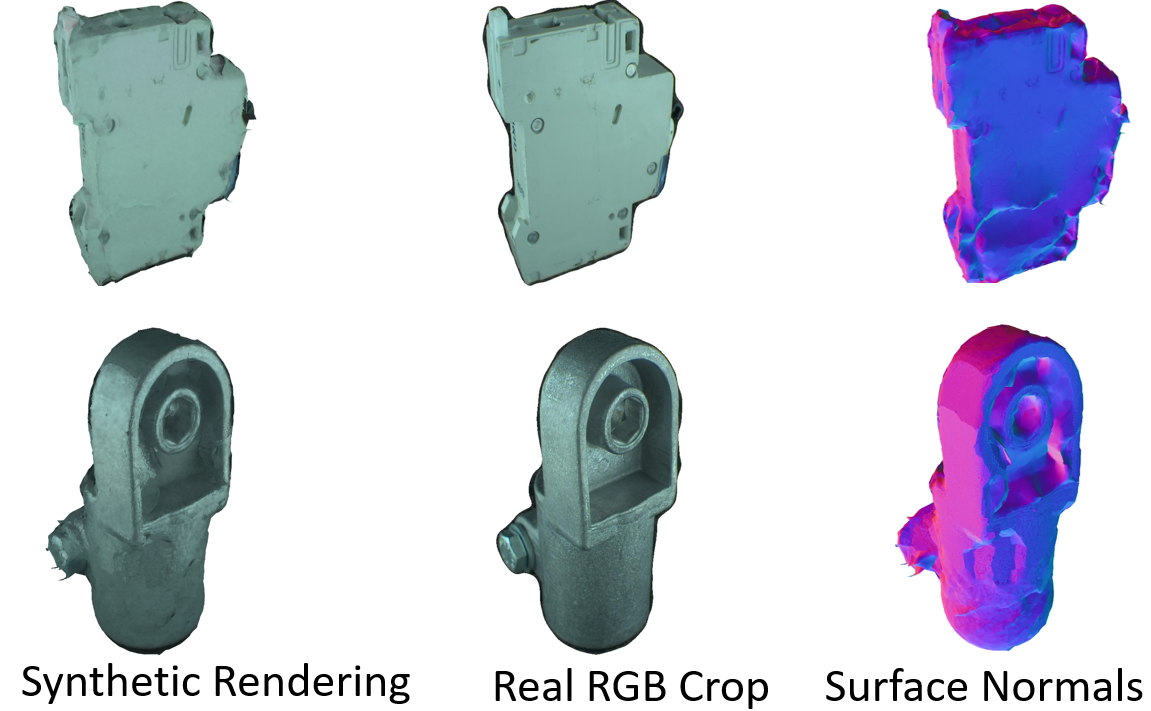}
    \caption{\textbf{Good NVDIFFRECMC Results}}
    \label{fig:DAfusionslight}
\end{subfigure}
\hfill
\begin{subfigure}[b]{0.49\textwidth}
    \centering
    \includegraphics[width=0.9\linewidth]{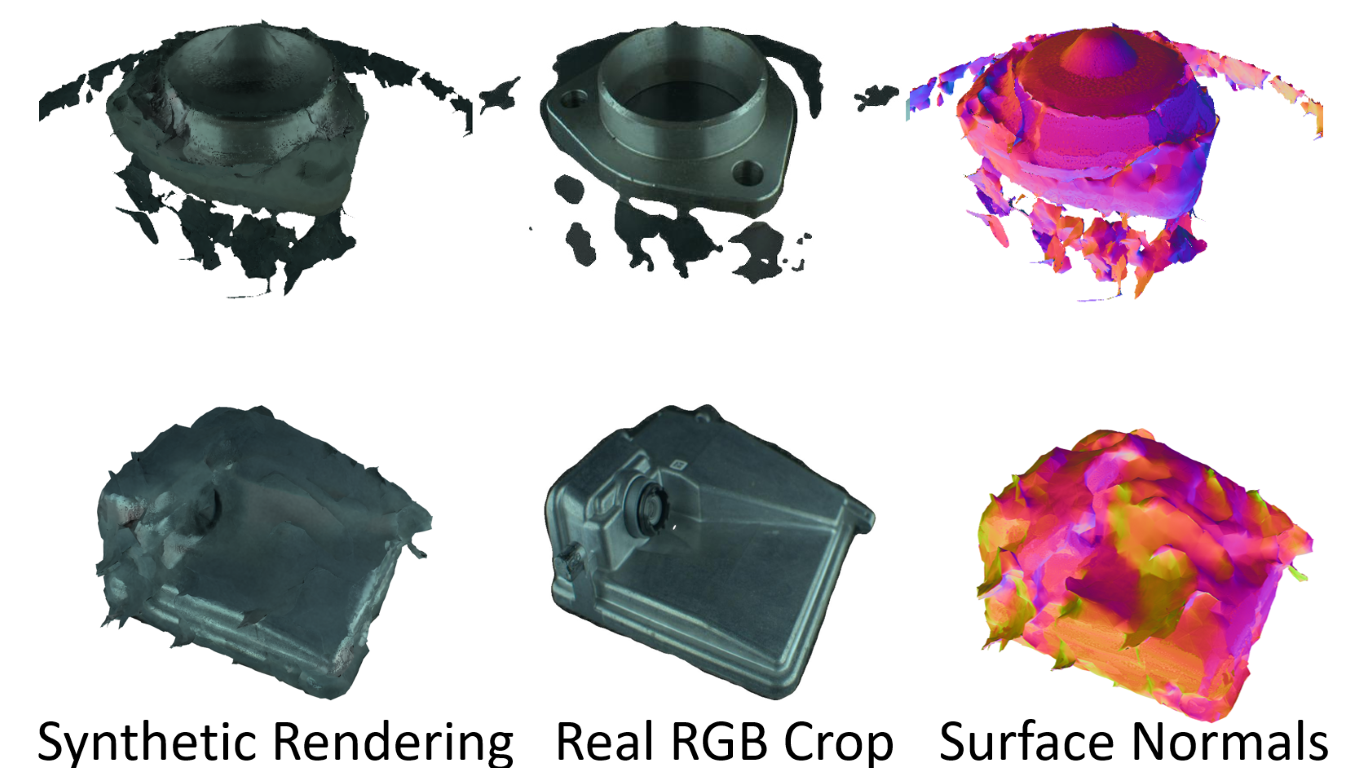}
    \caption{\textbf{Bad NVDIFFRECMC Results} }
    \label{fig:DAfusionsheavy}
\end{subfigure}
\caption{A)The textured CAD generation results in visually high fidelity. These textured CAD generations are well suited for synthetic data generation. B) In the top row the textured CAD generation fails due to bad generated masks of MVIP~\cite{MVIP}. The bottom row fails due to missing key features on the plain surface and thus faulty depth understanding.}
\label{fig:syntObject}
\end{figure}

For each object, NVDIFFRECMV is trained to produce the necessary assets, which are subsequently integrated into a Blender-based simulation pipeline. This pipeline is capable of rendering novel images from any perspective with simulated or randomized camera settings and lighting conditions. Due to the occlusion of the bottom part, objects must be synthesized multiple times to ensure comprehensive coverage. Our image capture uses a turntable with a fixed camera. However, the process can be further automated and enhanced using a robotic arm, eliminating the need for precise centering of the object on the turntable. We present qualitative results on object synthesizing in Fig.~\ref{fig:syntObject}. Here we observe that NVDIFFRECMV has difficulty to reconstruct correctly when A) provided with imprecise object masking in the input images, or B) the object has plain surfaces without key features, which are important to infer object shape from an imprecise depth understanding.

\textbf{Discussion on Synthesizing GenAI:} This method has significant potential, similar to previous methods of personalization and augmentation, as it can enhance the diversity of synthesized object texture files. Furthermore, it supports 3D perception applications, enabling automatic 3D augmentations such as bending or annotated 3D anomaly and defect generation on object surfaces. However, our investigations reveal limitations with respect to industrial applications. Despite addressing object masking effectively, the technology struggles with plain, featureless surfaces, a prevalent characteristic in industrial settings. This limitation impedes the use of NVDIFFRECMV in automated data generation pipelines due to a lack of robustness.

\section{Results}
In this section, we present two quantitative experiments designed to evaluate the benefits of automated synthetic data generation within the MVIP use-case.

\textbf{Experimental Setup:} We used a ResNet18~\cite{Resnet} image encoder without pre-training weights to isolate the impact of the generated data. The experiments were conducted on a single NVIDIA 3090 GPU. Each generated dataset extension for MVIP is partitioned into training and testing (20\%) datasets.

\textbf{Experiment 1:} We employ DA-Fusion~\cite{DAFusion} for image augmentation, creating two additional datasets for MVIP. We focus on 23 MVIP objects from the “CarComponent” super-class, which includes car generators and starters. This selection is crucial for addressing the similar challenge of MVIP's objects, visualized in Fig.~\ref{fig:mvipObjects}. The first dataset addition is an in-distribution (ID) dataset created using DA-Fusion with a low intensity factor. The second is a near out-of-distribution (near-OoD) dataset generated with high intensity image augmentation. Image personalization was excluded due to its similarity in purpose to the augmentation-based near-OoD dataset.

\begin{table}
    \centering
    \caption{The ablation results of in-distribution (ID) and out-of-distribution (OoD) generated dataset extensions to MVIP. Notice that OoD is not beneficial with respect to classification accuracy or confidence. However, OoD helps to drastically reduce overconfidence baseline on the test data. In contrast, ID data yields the best results and has the largest confidence gap between the ID and OoD test data. This gap is important indicate to users that the model is uncertain.}
    \begin{tabular}{l |c | c |c | c}
    \toprule
    Dataset & Top-1 Accurancy & avg. ID Conf. & avg. near-OoD Conf.& Conf. $\Delta$\\
    \midrule
    MVIP (baseline)&  71.4\% & 69\% & 62\% &7\%\\
    + ID Data & \textbf{77.5\%} & \textbf{76\%} & 65\% &\textbf{11\%}\\
    + OoD Data &  75.3\% & 40\% & \textbf{35\%} &5\%\\
    \bottomrule
    \end{tabular}
    \label{tab:learningResults}
\end{table}

\begin{figure}
\centering
\begin{subfigure}[t]{0.32\textwidth}
    \centering
    \includegraphics[width=1.0\linewidth]{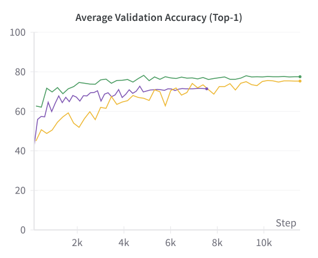}
    %\caption{A) }
    %\label{fig:TrainAcc}
\end{subfigure}
\begin{subfigure}[t]{0.32\textwidth}
    \centering
    \includegraphics[width=1.0\linewidth]{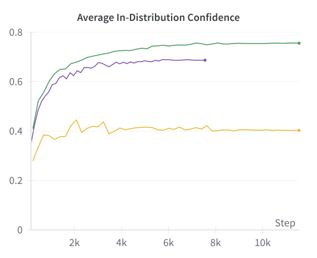}
    %\caption{B) }
    %\label{fig:TrainIOD}
\end{subfigure}
\begin{subfigure}[t]{0.32\textwidth}
    \centering
    \includegraphics[width=1.0\linewidth]{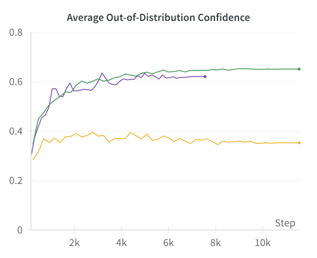}
    %\caption{B) }
    %\label{fig:TrainOOD}
\end{subfigure}
\caption{\textcolor{Plum}{MVIP (Baseline)}, \textcolor{Green}{+ID Data}, \textcolor{YellowOrange}{+ OoD Data}:  Training behavior of the ablation study for the supervised classification layer on top of the frozen pretrained (supervised contrastive learning~\cite{SuperContrast}) image encoder (ResNet18~\cite{Resnet}). We stop training after reaching a plateau.}
\label{fig:TrainCurvesDA}
\end{figure}
We conducted an ablation study against the baseline performance in MVIP. For each ablation iteration, the ResNet encoder is first pretrained using supervised contrastive learning~\cite{SuperContrast}. Subsequently, the encoder is frozen, and the final fully connected layer is finetuned with supervised learning on the labeled data. The supervised training behavior for each ablation iteration is illustrated in Fig.~\ref{fig:TrainCurvesDA}. Notice that the OoD data substantially reduce the model confidence while maintaining relative classification accuracy. The final results are detailed in Table~\ref{tab:learningResults}.

\begin{figure}
    \centering
    \begin{subfigure}[t]{0.58\textwidth}
    \centering
    \includegraphics[width=1.0\linewidth]{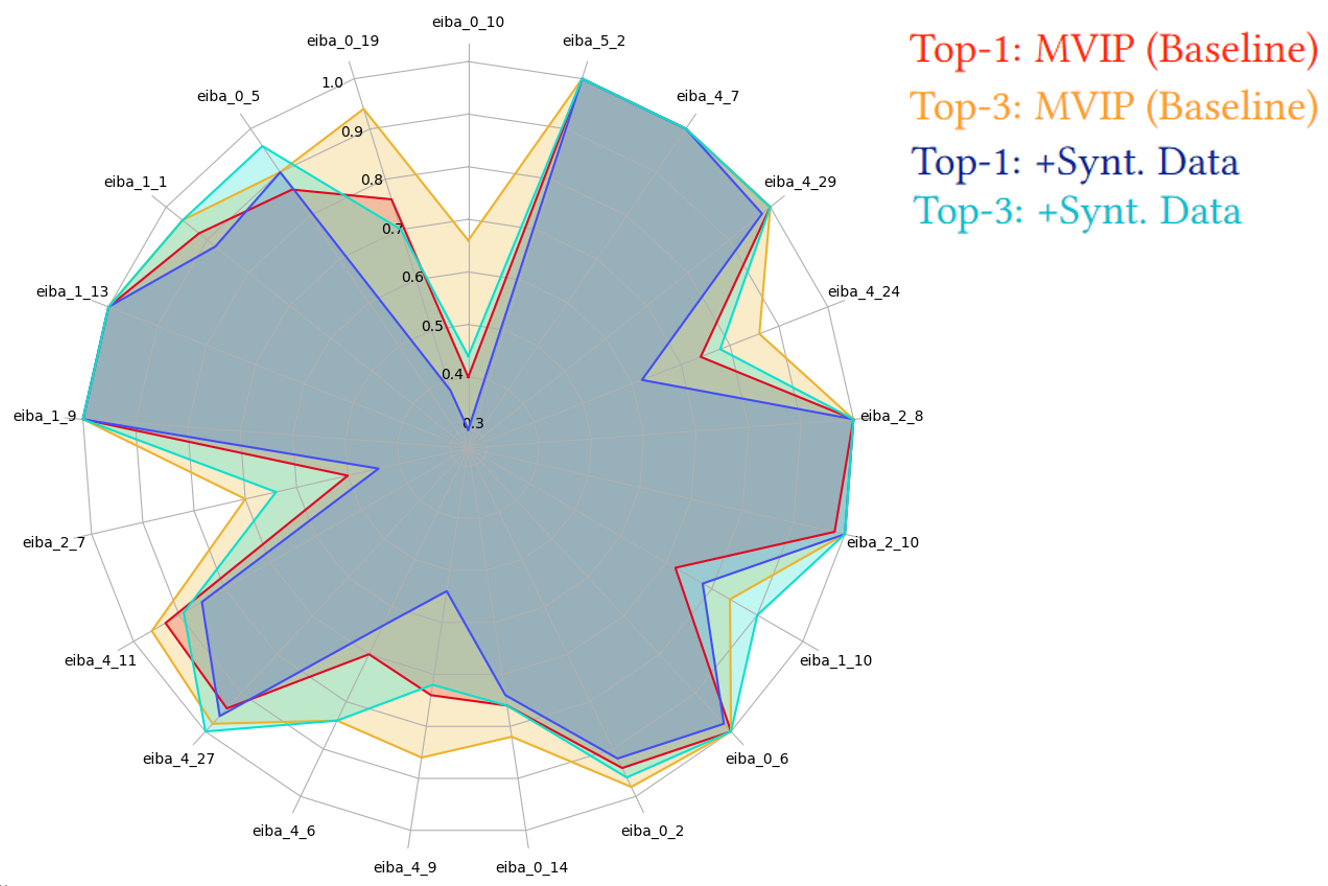}
    \caption{MVIP Object Results}
    %\label{fig:TrainIOD}
\end{subfigure}
\begin{subfigure}[t]{0.38\textwidth}
    \centering
    \includegraphics[width=1.0\linewidth]{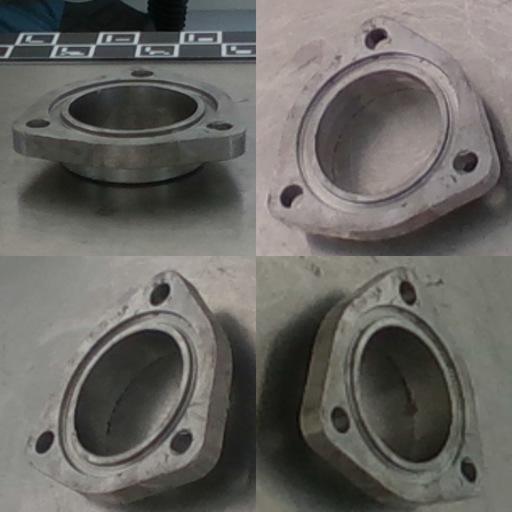}
    \caption{Object Class “eiba\_5\_0”}
    %\caption{B) }
    %\label{fig:TrainOOD}
\end{subfigure}
    \caption{Per Class Effect of using CAD-based synthetic (synt.) data comming form synthetized objects via NVDIFFRECMC~\cite{nvdiffrec}. We notice that the classification is affected when the synt. data is bad (see e.g.~\ref{fig:syntObject}). For the object left (b) one can see many features stains on the object surface, which helps to guide NVDIFFRECMC in precise synchronization.}
    \label{fig:MVIP_Diffrec}
\end{figure}

\textbf{Experiment 2:} We utilize NVDIFFRECMC~\cite{nvdiffrecmc} along with our Blender-based simulation pipeline to generate an in-distribution dataset extension for MVIP. This involves 21 physically available objects representing a cross-section of MVIP's object diversity to identify key features where NVDIFFRECMC fails (examples shown in Fig.~\ref{fig:mvipObjects}). This extension of the dataset is then used for supervised classification training using ResNet18~\cite{Resnet} (see the results in  Fig.~\ref{fig:MVIP_Diffrec})

\section{Conclusion}
This paper presents an extensive review of GenAI-based methods for data generation and augmentation in the context of the industrial computer vision application raised by MVIP~\cite{MVIP}. Our investigation highlights the potential of these methods to enrich training datasets, thereby improving model performance in challenging scenarios such as the high object similarity in MVIP. The experiments demonstrate that GenAI techniques can effectively diversify data distributions, enhance feature extraction, and address the “chicken-and-egg” dilemma of extensive dataset creation and model development. Despite these improvements, challenges remain, particularly in automating data generation pipelines while ensuring robust and reliable performance across varied industrial settings. Our qualitative insights into current limitations are valuable for guiding related work in the continuous refinement and adaptation of GenAI methods to fully exploit their potential in real-world applications. To ensure reproducibility, we will upload and share collected data, generated data, and code on our github: \textcolor{pink}{\url{https://github.com/tbd}}

\section{Broader Impact Statement}
% The 9 pages allocated for the main paper must include a broader impact
% statement regarding the approach, datasets and applications proposed/used in
% your paper. It should reflect on the environmental, ethical and societal
% implications of your work. The statement should require at most one page and
% must be included both at submission and camera-ready time.
%
% If authors have reflected on their work and determined that there are no
% likely negative broader impacts, they may use the following statement:
%
% After careful reflection, the authors have determined that this work presents
% no notable negative impacts to society or the environment.
%
% This section is included in the template as a default, but you can also place these
% discussions anywhere else in the main paper, e.g., in the introduction/future work.
%
% The Centre for the Governance of AI has written an excellent guide for writing
% good broader impact statements (for the NeurIPS conference) that may be a
% useful resource for AutoML-Conf authors:
%
% https://medium.com/@GovAI/a-guide-to-writing-the-neurips-impact-statement-4293b723f832
The integration of GenAI-based data generation and augmentation methods in industrial applications offers significant benefits, such as enhanced model predictability and efficiency in data acquisition -- a key driving cost factor. These improvements can foster greater trust in AI technologies, potentially accelerating their adoption across various sectors. However, ethical considerations must be addressed about the potential biases inherent in generated datasets, and job security in the industry -- thus driving AI-assistance not only full automation. In addition, environmental impacts related to the computational resources required to train complex models should be carefully managed. By fostering collaboration between industry experts and AI researchers, we can ensure that these technologies are developed responsibly, maximizing their positive impact on society while mitigating potential adverse effects.

After careful reflection, the authors have determined that this work does not have significant negative impacts on society or the environment. However, continued vigilance and assessment of GenAI methods are crucial as they evolve and are deployed in broader applications.


\bibliographystyle{abbrvnat}
\bibliography{bibliography}

\end{document}